# MobileFaceNets: Efficient CNNs for Accurate Real-Time Face Verification on Mobile Devices

Sheng Chen[1,2], Yang Liu[2], Xiang Gao[2], and Zhen Han[1]

[1] School of Computer and Information Technology, Beijing Jiaotong University, Beijing, China
[2] Research Institute, Watchdata Inc., Beijing, China
{sheng.chen, yang.liu.yj, xiang.gao}@watchdata.com,
zhan@bjtu.edu.cn

**Abstract.** We present a class of extremely efficient CNN models, MobileFaceNets, which use less than 1 million parameters and are specifically tailored for high-accuracy real-time face verification on mobile and embedded devices. We first make a simple analysis on the weakness of common mobile networks for face verification. The weakness has been well overcome by our specifically designed MobileFaceNets. Under the same experimental conditions, our MobileFaceNets achieve significantly superior accuracy as well as more than 2 times actual speedup over MobileNetV2. After trained by ArcFace loss on the refined MS-Celeb-1M, our single MobileFaceNet of 4.0MB size achieves 99.55% accuracy on LFW and 92.59% TAR@FAR1e-6 on MegaFace, which is even comparable to state-of-the-art big CNN models of hundreds MB size. The fastest one of MobileFaceNets has an actual inference time of 18 milliseconds on a mobile phone. For face verification, MobileFaceNets achieve significantly improved efficiency over previous state-of-the-art mobile CNNs.

**Keywords:** Mobile network, face verification, face recognition, convolutional neural network, deep learning.

## 1 Introduction

Face verification is an important identity authentication technology used in more and more mobile and embedded applications such as device unlock, application login, mobile payment and so on. Some mobile applications equipped with face verification technology, for example, smartphone unlock, need to run offline. To achieve user-friendliness with limited computation resources, the face verification models deployed locally on mobile devices are expected to be not only accurate but also small and fast. However, modern high-accuracy face verification models are built upon deep and big convolutional neural networks (CNNs) which are supervised by novel loss functions during training stage. The big CNN models requiring high computational resources are not suitable for many mobile and embedded applications. Several highly efficient neural network architectures, for example, MobileNetV1 [1],

ShuffleNet [2], and MobileNetV2 [3], have been proposed for common visual recognition tasks rather than face verification in recent years. It is a straight-forward way to use these common CNNs unchanged for face verification, which only achieves very inferior accuracy compared with state-of-the-art results according to our experiments (see Table 2).

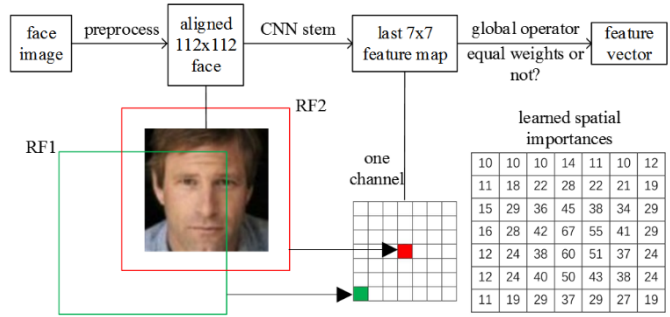

**Fig. 1.** A typical face feature embedding CNN and the receptive field (RF). The last 7x7 feature map is denoted as FMap-end. RF1 and RF2 correspond to the corner unit and the center unit in FMap-end respectively. The corner unit should be of less importance than the center unit. When a global depthwise convolution (GDConv) is used as the global operator, for a fixed spatial position, the norm of the weight vector consisted of GDConv weights in all channels can be considered as the spatial importance. We show that GDConv learns very different importances at different spatial positions after training.

In this paper, we make a simple analysis on common mobile networks' weakness for face verification. The weakness has been well overcome by our specifically designed MobileFaceNets, which is a class of extremely efficient CNN models tailored for high-accuracy real-time face verification on mobile and embedded devices. Our MobileFaceNets use less than 1 million parameters. Under the same experimental conditions, our MobileFaceNets achieve significantly superior accuracy as well as more than 2 times actual speedup over MobileNetV2. After trained on the refined MS-Celeb-1M [4] by ArcFace [5] loss from scratch, our single MobileFaceNet model of 4.0MB size achieves 99.55% face verification accuracy (see Table 3) on LFW [6] and 92.59% TAR@FAR10-6 (see Table 4) on MegaFace Challenge 1 [7], which is even comparable to state-of-the-art big CNN models of hundreds MB size. Note that many existing techniques such as pruning [37], low-bit quantization [29], and knowledge distillation [16] are able to improve MobileFaceNets' efficiency additionally, but these are not included in the scope of this paper.

The major contributions of this paper are summarized as follows: (1) After the last (non-global) convolutional layer of a face feature embedding CNN, we use a global depthwise convolution layer rather than a global average pooling layer or a fully connected layer to output a discriminative feature vector. The advantage of this choice is also analyzed in both theory and experiment. (2) We carefully design a class of face feature embedding CNNs, namely MobileFaceNets, with extreme efficiency on mobile and embedded devices. (3) Our experiments on LFW, AgeDB ([8]), and

MegaFace show that our MobileFaceNets achieve significantly improved efficiency over previous state-of-the-art mobile CNNs for face verification.

## 2 Related Work

Tuning deep neural architectures to strike an optimal balance between accuracy and performance has been an area of active research for the last several years [3]. For common visual recognition tasks, many efficient architectures have been proposed recently [1, 2, 3, 9]. Some efficient architectures can be trained from scratch. For example, SqueezeNet ([9]) uses a bottleneck approach to design a very small network and achieves AlexNet-level [10] accuracy on ImageNet [11, 12] with 50x fewer parameters (i.e., 1.25 million). MobileNetV1 [1] uses depthwise separable convolutions to build lightweight deep neural networks, one of which, i.e., MobileNet-160 (0.5x), achieves 4% better accuracy on ImageNet than SqueezeNet at about the same size. ShuffleNet [2] utilizes pointwise group convolution and channel shuffle operation to reduce computation cost and achieve higher efficiency than MobileNetV1. MobileNetV2 [3] architecture is based on an inverted residual structure with linear bottleneck and improves the state-of-the-art performance of mobile models on multiple tasks and benchmarks. The mobile NASNet [13] model, which is an architectural search result with reinforcement learning, has much more complex structure and much more actual inference time on mobile devices than MobileNetV1, ShuffleNet, and MobileNetV2. However, these lightweight basic architectures are not so accurate for face verification when trained from scratch (see Table 2).

Accurate lightweight architectures specifically designed for face verification have been rarely researched. [14] presents a light CNN framework to learn a compact embedding on the large-scale face data, in which the Light CNN-29 model achieves 99.33% face verification accuracy on LFW with 12.6 million parameters. Compared with MobileNetV1, Light CNN-29 is not lightweight for mobile and embedded platform. Light CNN-4 and Light CNN-9 are much less accurate than Light CNN-29. [15] proposes ShiftFaceNet based on ShiftNet-C model with 0.78 million parameters, which only achieves 96.0% face verification accuracy on LFW. In [5], an improved version of MobileNetV1, namely LMobileNetE, achieves comparable face verification accuracy to state-of-the-art big models. But LMobileNetE is actually a big model of 112MB model size, rather than a lightweight model. All above models are trained from scratch.

Another approach for obtaining lightweight face verification models is compressing pretrained networks by knowledge distillation [16]. In [17], a compact student network (denoted as MobileID) trained by distilling knowledge from the teacher network DeepID2+ [33] achieves 97.32% accuracy on LFW with 4.0MB model size. In [1], several small MobileNetV1 models for face verification are trained by distilling knowledge from the pretrained FaceNet [18] model and only face verification accuracy on the authors' private test dataset are reported. Regardless of the small student models' accuracy on public test datasets, our MobileFaceNets achieve comparable accuracy to the strong teacher model FaceNet on LFW (see Table 3) and MegaFace (see Table 4).

# 3 Approach

In this section, we will describe our approach towards extremely efficient CNN models for accurate real-time face verification on mobile devices, which overcome the weakness of common mobile networks for face verification. To make our results totally reproducible, we use ArcFace loss to train all face verification models on public datasets, following the experimental settings in [5].

## 3. 1  The Weakness of Common Mobile Networks for Face Verification

There is a global average pooling layer in most recent state-of-the-art mobile networks proposed for common visual recognition tasks, for example, MobileNetV1, ShuffleNet, and MobileNetV2. For face verification and recognition, some researchers ([14], [5], etc.) have observed that CNNs with global average pooling layers are less accurate than those without global average pooling. However, no theoretical analysis for this phenomenon has been given. Here we make a simple analysis on this phenomenon in the theory of receptive field [19].

A typical deep face verification pipeline includes preprocessing face images, extracting face features by a trained deep model, and matching two faces by their features' similarity or distance. Following the preprocessing method in [5, 20, 21, 22], we use MTCNN [23] to detect faces and five facial landmarks in images. Then we align the faces by similarity transformation according to the five landmarks. The aligned face images are of size 112 × 112, and each pixel in RGB images is normalized by subtracting 127.5 then divided by 128. Finally, a face feature embedding CNN maps each aligned face to a feature vector, as shown in Fig. 1. Without loss of generality, we use MobileNetV2 as the face feature embedding CNN in the following discussion. To preserve the same output feature map sizes as the original network with 224 × 224 input, we use the setting of stride = 1 in the first convolutional layer instead of stride = 2, where the latter setting leads to very poor accuracy. So, before the global average pooling layer, the output feature map of the last convolutional layer, denoted as FMap-end for convenience, is of spatial resolution 7 × 7. Although the theoretical receptive fields of the corner units and the central units of FMap-end are of the same size, they are at different positions of the input image. The receptive fields' center of FMap-end's corner units is in the A typical deep face verification pipeline includes preprocessing face images, extracting face features by a trained deep model, and matching two faces by their features' similarity or distance. Following the preprocessing method in [5, 20, 21, 22], we use MTCNN [23] to detect faces and five facial landmarks in images. Then we align the faces by similarity transformation according to the five landmarks. The aligned face images are of size 112 × 112, and each pixel in RGB images is normalized by subtracting 127.5 then divided by 128. Finally, a face feature embedding CNN maps each aligned face to a feature vector, as shown in Fig. 1. Without loss of generality, we use MobileNetV2 as the face feature embedding CNN in the following discussion. To preserve the same output feature map sizes as the original network with 224 × 224 input, we use the setting of stride = 1 in the first convolutional layer instead of stride = 2, where the latter setting leads to very poor accuracy. So, before the global average pooling layer, the output feature map of the last convolutional layer, denoted as FMap-end for convenience, is of spatial resolution 7 × 7. Although the theoretical receptive fields of the corner units and the central units of FMap-end are of the same size, they are at different positions of the input image. The receptive fields' center of FMap-end's corner units is in the corner of the input image and the receptive fields' center of FMend's central units are in the center of the input image, as shown in Fig. 1. According to [24], pixels at the center of a receptive field have a much larger impact on an output and the distribution of impact within a receptive field on the output is nearly Gaussian. The effective receptive field [24] sizes of FMap-end's corner units are much smaller than the ones of FMap-end's central units. When the input image is an aligned face, a corner unit of FMap-end carries less information of the face than a central unit. Therefore, different units of FMap-end are of different importance for extracting a face feature vector.

In MobileNetV2, the flattened FMap-end is unsuitable to be directly used as a face feature vector since it is of a too high dimension 62720. It is a natural choice to use the output of the global average pooling (denoted as GAPool) layer as a face feature vector, which achieves inferior verification accuracy in many researchers' experiments [14, 5] as well as ours (see Table 2). The global average pooling layer treats all units of FMap-end with equal importance, which is unreasonable according to the above analysis. Another popular choice is to replace the global average pooling layer with a fully connected layer to project FMap-end to a compact face feature vector, which adds large number of parameters to the whole model. Even when the face feature vector is of a low dimension 128, the fully connected layer after FMap-end will bring additional 8 million parameters to MobileNetV2. We do not consider this choice since small model size is one of our pursuits.

### 3.2 Global Depthwise Convolution

To treat different units of FMap-end with different importance, we replace the global average pooling layer with a global depthwise convolution layer (denoted as GDConv). A GDConv layer is a depthwise convolution (c.f. [25, 1]) layer with kernel size equaling the input size, pad = 0, and stride = 1. The output for global depthwise convolution layer is computed as:

$$G_m = \sum_{i,j} K_{i,j,m} \cdot F_{i,j,m} \quad (1)$$

where $F$ is the input feature map of size $W \times H \times M$, $K$ is the depthwise convolution kernel of size $W \times H \times M$, $G$ is the output of size $1 \times 1 \times M$, the $m_{th}$ channel in $G$ has only one element $G_m$, $(i,j)$ denotes the spatial position in $F$ and $K$, and $m$ denotes the channel index.

Global depthwise convolution has a computational cost of:

$$W \cdot H \cdot M \quad (2)$$

When used after FMap-end in MobileNetV2 for face feature embedding, the global depthwise convolution layer of kernel size 7 × 7 × 1280 outputs a 1280-dimensional face feature vector with a computational cost of 62720 MAdds (i.e., the number of operations measured by multiply-adds, c.f. [3]) and 62720 parameters. Let MobileNetV2-GDConv denote MobileNetV2 with global depthwise convolution layer. When both MobileNetV2 and MobileNetV2-GDConv are trained on CIASIA-Webface [26] for face verification by ArcFace loss, the latter achieves significantly better accuracy on LFW and AgeDB (see Table 2). Global depthwise convolution layer is an efficient structure for our design of MobileFaceNets.

**Table 1.** MobileFaceNet architecture for feature embedding. We use almost the same notations as MobileNetV2 [3]. Each line describes a sequence of operators, repeated $n$ times. All layers in the same sequence have the same number $c$ of output channels. The first layer of each sequence has a stride $s$ and all others use stride 1. All spatial convolutions in the bottlenecks use $3 \times 3$ kernels. The expansion factor $t$ is always applied to the input size. GDConv7x7 denotes GDConv of $7 \times 7$ kernels.

| Input | Operator | $t$ | $c$ | $n$ | $s$ |
|---|---|---|---|---|---|
| $112^2 \times 3$ | conv3x3 | - | 64 | 1 | 2 |
| $56^2 \times 64$ | depthwise conv3x3 | - | 64 | 1 | 1 |
| $56^2 \times 64$ | bottleneck | 2 | 64 | 5 | 2 |
| $28^2 \times 64$ | bottleneck | 4 | 128 | 1 | 2 |
| $14^2 \times 128$ | bottleneck | 2 | 128 | 6 | 1 |
| $14^2 \times 128$ | bottleneck | 4 | 128 | 1 | 2 |
| $7^2 \times 128$ | bottleneck | 2 | 128 | 2 | 1 |
| $7^2 \times 128$ | conv1x1 | - | 512 | 1 | 1 |
| $7^2 \times 512$ | linear GDConv7x7 | - | 512 | 1 | 1 |
| $1^2 \times 512$ | linear conv1x1 | - | 128 | 1 | 1 |

### 3.3 MobileFaceNet Architectures

Now we describe our MobileFaceNet architectures in detail. The residual [38] bottlenecks proposed in MobileNetV2 [3] are used as our main building blocks. For convenience, we use the same conceptions as those in [3]. The detailed structure of our primary MobileFaceNet architecture is shown in Table 1. Particularly, expansion factors for bottlenecks in our architecture are much smaller than those in MobileNetV2. We use PReLU [27] as the non-linearity, which is slightly better for face verification than using ReLU (see Table 2). In addition, we use a fast downsampling strategy at the beginning of our network, an early dimension-reduction strategy at the last several convolutional layers, and a linear $1 \times 1$ convolution layer following a linear global depthwise convolution layer as the feature output layer. Batch normalization [28] is utilized during training and batch normalization folding (c.f. Section 3.2 of [29]) is applied before deploying.

Our primary MobileFaceNet network has a computational cost of 221 million MAdds and uses 0.99 million parameters. We further tailor our primary architecture as follows. To reduce computational cost, we change input resolution from $112 \times 112$ to $112 \times 96$ or $96 \times 96$. To reduce the number of parameters, we remove the linear $1 \times 1$ convolution layer after the linear GDConv layer from MobileFaceNet, the resulting network of which is called MobileFaceNet-M. From MobileFaceNet-M, removing the

1 × 1 convolution layer before the linear GDConv layer produces the smallest network called MobileFaceNet-S. These MobileFaceNet networks' effectiveness is demonstrated by the experiments in the next section.

## 4 Experiments

In this section, we will first describe the training settings of our MobileFaceNet models and our baseline models. Then we will compare the performance of our trained face verification models with some previous published face verification models, including several state-of-the-art big models.

### 4.1 Training settings and accuracy comparison on LFW and AgeDB

We use MobileNetV1, ShuffleNet, and MobileNetV2 (with stride = 1 for the first convolutional layers of them since the setting of stride = 2 leads to very poor accuracy) as our baseline models. All MobileFaceNet models and baseline models are trained on CASIA-Webface dataset from scratch by ArcFace loss, for a fair performance comparison among them. We set the weight decay parameter to be 4e-5, except the weight decay parameter of the last layers after the global operator (GDConv or GAPool) being 4e-4. We use SGD with momentum 0.9 to optimize models and the batch size is 512. The learning rate begins with 0.1 and is divided by 10 at the 36K, 52K and 58K iterations. The training is finished at 60K iterations. Then, the face verification accuracy on LFW and AgeDB-30 is compared in Table 2.

**Table 2.** Performance comparison among mobile models trained on CASIA-Webface. In the last column, we report actual inference time in milliseconds (ms) on a Qualcomm Snapdragon 820 CPU of a mobile phone with 4 threads (using NCNN [30] inference framework).

| Network | LFW | AgeDB-30 | Params | Speed |
|---|---|---|---|---|
| MobileNetV1 | 98.63% | 88.95% | 3.2M | 60ms |
| ShuffleNet (1×, g = 3) | 98.70% | 89.27% | **0.83M** | 27ms |
| MobileNetV2 | 98.58% | 88.81% | 2.1M | 49ms |
| MobileNetV2-GDConv | 98.88% | 90.67% | 2.1M | 50ms |
| **MobileFaceNet** | **99.28%** | **93.05%** | 0.99M | 24ms |
| MobileFaceNet (112 × 96) | 99.18% | 92.96% | 0.99M | 21ms |
| MobileFaceNet (96 × 96) | 99.08% | 92.63% | 0.99M | **18ms** |
| MobileFaceNet-M | 99.18% | 92.67% | 0.92M | 24ms |
| MobileFaceNet-S | 99.00% | 92.48% | **0.84M** | 23ms |
| MobileFaceNet (ReLU) | 99.15% | 92.83% | 0.98M | 23ms |
| MobileFaceNet (expansion factor ×2) | 99.10% | 92.81% | 1.1M | 27ms |

As shown in Table 2, compared with the baseline models of common mobile networks, our MobileFaceNets achieve significantly better accuracy with faster inference speed. Our primary MobileFaceNet achieves the best accuracy and MobileFaceNet with a lower input resolution of 96 × 96 has the fastest inference speed. Note that our MobileFaceNets are more efficient than those with larger expansion factor such as MobileFaceNet (expansion factor ×2) and MobileNetV2-GDConv.

To pursue ultimate performance, MobileFaceNet, MobileFaceNet (112 × 96), and MobileFaceNet (96 × 96) are also trained by ArcFace loss on the cleaned training set of MS-Celeb-1M database [5] with 3.8M images from 85K subjects. The accuracy of our primary MobileFaceNet is boosted to 99.55% and 96.07% on LFW and AgeDB-30, respectively. The three trained models' accuracy on LFW is compared with previous published face verification models in Table 3.

**Table 3.** Performance comparison with previous published face verification models on LFW.

| Method | Training Data | #Net | Model Size | LFW Acc. |
|---|---|---|---|---|
| Deep Face [31] | 4M | 3 | - | 97.35% |
| DeepFR [32] | 2.6M | 1 | 0.5GB | 98.95% |
| DeepID2+ [33] | 0.3M | 25 | - | 99.47% |
| Center Face [34] | 0.7M | 1 | 105MB | 99.28% |
| DCFL [35] | 4.7M | 1 | - | 99.55% |
| SphereFace [20] | 0.49M | 1 | - | 99.47% |
| CosFace [22] | 5M | 1 | - | 99.73% |
| ArcFace (LResNet100E-IR) [5] | 3.8M | 1 | 250MB | **99.83%** |
| FaceNet [18] | 200M | 1 | 30MB | 99.63% |
| ArcFace (LMobileNetE) [5] | 3.8M | 1 | 112MB | 99.50% |
| Light CNN-29 [14] | 4M | 1 | 50MB | 99.33% |
| MobileID [17] | - | 1 | 4.0MB | 97.32% |
| ShiftFaceNet [15] | - | 1 | **3.1MB** | 96.00% |
| **MobileFaceNet** | 3.8M | 1 | **4.0MB** | 99.55% |
| MobileFaceNet (112 × 96) | 3.8M | 1 | **4.0MB** | 99.53% |
| MobileFaceNet (96 × 96) | 3.8M | 1 | **4.0MB** | 99.52% |

### 4.2  Evaluation on MegaFace Challenge1

**Table 4.** Face verification evaluation on Megafce Challenge 1. "VR" refers to face verification TAR (True Accepted Rate) under $10^{-6}$ FAR (False Accepted Rate). MobileFaceNet (R) are evaluated on the refined version of MegaFace dataset (c.f. [5]).

| Method | VR (large protocol) | VR (small protocol) |
|---|---|---|
| SIAT MMLAB [34] | 87.27% | 76.72% |
| DeepSense V2 | 95.99% | 82.85% |
| SphereFace-Small [20] | - | 90.04% |
| Google-FaceNet v8 [18] | 86.47% | - |
| Vocord-deepVo V3 | 94.96% | - |
| CosFace (3-patch) [22] | 97.96% | **92.22%** |
| iBUG_DeepInsight (ArcFace [5]) | **98.48%** | - |
| **MobileFaceNet** | 90.16% | 85.76% |
| **MobileFaceNet (R)** | 92.59% | 88.09% |

In this paper, we use the Facescrub [36] dataset as the probe set to evaluate the verification performance of our primary MobileFaceNet on Megaface Challenge 1. Table 4 summarizes the results of our models trained on two protocols of MegaFace where the training dataset is regarded as small if it has less than 0.5 million images, large otherwise. Our primary MobileFaceNet shows comparable accuracy for the verification task on both the protocols.

## 5 Conclusion

We proposed a class of face feature embedding CNNs, namely MobileFaceNets, with extreme efficiency for real-time face verification on mobile and embedded devices. Our experiments show that MobileFaceNets achieve significantly improved efficiency over previous state-of-the-art mobile CNNs for face verification.

**Acknowledgments.** We thank Jia Guo for helpful discussion, and thank Yang Wang, Lian Li, Licang Qin, Yan Gao, Hua Chen, and Min Zhao for application development.

## References

1. Howard, A. G., Zhu, M., Chen, B., Kalenichenko, D., Wang, W., Weyand, T., et al.: Mobilenets: Efficient convolutional neural networks for mobile vision applications. CoRR, abs/1704.04861 (2017)
2. Zhang, X., Zhou, X., Lin, M., Sun, J.: Shufflenet: An extremely efficient convolutional neural network for mobile devices. CoRR, abs/1707.01083 (2017)
3. Sandler, M., Howard, A., Zhu, M., Zhmoginov, A., Chen, L.C.: MobileNetV2: Inverted Residuals and Linear Bottlenecks. CoRR, abs/1801.04381 (2018)
4. Guo, Y., Zhang, L., Hu, Y., He, X., Gao, J.: Ms-celeb-1m: A dataset and benchmark for large-scale face recognition. arXiv preprint, arXiv: 1607.08221 (2016)
5. Deng, J., Guo, J., Zafeiriou, S.: ArcFace: Additive Angular Margin Loss for Deep Face Recognition. arXiv preprint, arXiv: 1801.07698 (2018)
6. Huang, G.B., Ramesh, M., Berg, T., et al.: Labeled faces in the wild: a database for studying face recognition in unconstrained environments. (2007)
7. Kemelmacher-Shlizerman, I., Seitz, S. M., Miller, D., Brossard, E.: The megaface benchmark: 1 million faces for recognition at scale. In: CVPR (2016)
8. Moschoglou, S., Papaioannou, A., Sagonas, C., Deng, J., Kotsia, I., Zafeiriou, S.: Agedb: The first manually collected in-the-wild age database. In: CVPRW (2017)
9. Iandola, F. N., Han, S., Moskewicz, M.W., Ashraf, K., Dally, W.J., Keutzer, K.: Squeezenet: Alexnet-level accuracy with 50x fewer parameters and 0.5 mb model size. arXiv preprint, arXiv:1602.07360 (2016)
10. Krizhevsky, A., Sutskever, I., Hinton, G.E.: Imagenet classification with deep convolutional neural networks. In: NIPS (2012)
11. Deng, J., Dong, W., Socher, R., Li, L.J., Li, K., Fei-Fei, L.: ImageNet: a large-scale hierarchical image database. In: CVPR. IEEE (2009)
12. Russakovsky, O., Deng, J., Su, H., et al.: Imagenet large scale visual recognition challenge. Int. J. Comput. Vis. 115, 211–252 (2015)
13. Zoph, B., Vasudevan, V., Shlens, J., Le, Q.V.: Learning transferable architectures for scalable image recognition. arXiv preprint, arXiv:1707.07012 (2017)

14. Wu, X., He, R., Sun, Z., Tan, T.: A light cnn for deep face representation with noisy labels. arXiv preprint, arXiv:1511.02683 (2016)
15. Wu, B., Wan, A., Yue, X., Jin, P., Zhao, S., Golmant, N., et al.: Shift: A Zero FLOP, Zero Parameter Alternative to Spatial Convolutions. arXiv preprint, arXiv: 1711.08141 (2017)
16. Hinton, G. E., Vinyals, O., Dean, J.: Distilling the knowledge in a neural network. In arXiv:1503.02531 (2015)
17. Luo, P., Zhu, Z., Liu, Z., Wang, X., Tang, X., Luo, P., et al.: Face Model Compression by Distilling Knowledge from Neurons. In: AAAI (2016)
18. Schroff, F., Kalenichenko, D., Philbin, J.: Facenet: a unified embedding for face recognition and clustering. In: CVPR (2015)
19. Long, J., Zhang, N., Darrell, T.: Do convnets learn correspondence? Advances in Neural Information Processing Systems, 2, 1601-1609 (2014)
20. Liu, W., Wen, Y., Yu, Z., Li, M., Raj, B., Song, L.: Sphereface: Deep hypersphere embedding for face recognition. In: CVPR (2017)
21. Wang, F., Cheng, J., Liu, W., Liu, H.: Additive margin softmax for face verification. IEEE Signal Proc. Let., **25**(7), 926-930 (2018)
22. Wang, H., Wang, Y., Zhou, Z., Ji, X., Gong, D., Zhou, J., et al.: CosFace: Large Margin Cosine Loss for Deep Face Recognition. In arXiv: 1801.0941 (2018)
23. Zhang, K., Zhang, Z., Li, Z., Qiao, Y.: Joint Face Detection and Alignment using Multi-task Cascaded Convolutional Networks. IEEE Signal Proc. Let., **23**(10):1499–1503, 2016.
24. Luo, W., Li, Y., Urtasun, R., Zemel, R.: Understanding the Effective Receptive Field in Deep Convolutional Neural Networks. In: NIPS (2016)
25. Chollet, F.: Xception: Deep learning with depthwise separable convolutions. arXiv preprint, arXiv:1610.02357 (2016)
26. Yi, D., Lei, Z., Liao, S., Li, S. Z.: Learning face representation from scratch. arXiv preprint, arXiv:1411.7923 (2014)
27. He, K., Zhang, X., Ren, S., Sun, J.: Delving deep into rectifiers: Surpassing human-level performance on imagenet classification. In: CVPR (2015)
28. Ioffe, S., Szegedy, C.: Batch normalization: accelerating deep network training by reducing internal covariate shift. In: International Conference on Machine Learning (2015)
29. Jacob, B., Kligys, S., Chen, B., Zhu, M., Tang, M., Howard, A., et al.: Quantization and Training of Neural Networks for Efficient Integer-Arithmetic-Only Inference. arXiv preprint, arXiv: 1712.05877 (2017)
30. NCNN: a high-performance neural network inference framework optimized for the mobile platform, https://github.com/Tencent/ncnn, the version in Apr 20, 2018.
31. Taigman, Y., Yang, M., Ranzato, M., et al.: DeepFace: closing the gap to human-level performance in face verification. In: CVPR (2014)
32. Omkar M Parkhi, Andrea Vedaldi, Andrew Zisserman, et al, "Deep face recognition," In BMVC, volume 1, page 6, 2015.
33. Sun, Y., Wang, X., Tang, X.: Deeply learned face representations are sparse, selective, and robust. In: Computer Vision and Pattern Recognition, pp. 2892–2900 (2015).
34. Wen, Y., Zhang, K., Li, Z., Qiao, Y.: A discriminative feature learning approach for deep face recognition. In: ECCV (2016)
35. Deng, W., Chen, B., Fang, Y., Hu, J.: Deep Correlation Feature Learning for Face Verification in the Wild. IEEE Signal Proc. Let., **24**(12), 1877 – 1881 (2017)
36. Ng, H. W., Winkler, S.: A data-driven approach to cleaning large face datasets. In: IEEE International Conference on Image Processing (ICIP), pp. 343–347 (2014)
37. Han, S., Mao, H., Dally, W. J.: Deep compression: Compressing deep neural network with pruning, trained quantization and Huffman coding. CoRR, abs/1510.00149 (2015)
38. He, K., Zhang, X., Ren, S., Sun, J.: Deep residual learning for image recognition. In: CVPR (2016)